\documentclass[letterpaper, 10 pt, conference]{ieeeconf}  % Comment this line out if you need a4paper

\IEEEoverridecommandlockouts                              % This command is only needed if
\usepackage{graphics} % for pdf, bitmapped graphics files
\usepackage{amsmath} % assumes amsmath package installed
\usepackage{amssymb}  % assumes amsmath package installed
\usepackage{graphicx}
\usepackage{flushend}

\title{\LARGE \bf
On Training Flexible Robots using Deep Reinforcement Learning
}

\author{Zach Dwiel$^{1*}$ and Madhavun Candadai$^{2*}$ and Mariano Phielipp$^{3}$% <-this % stops a space
\thanks{$^{*}$with equal contributions}%
\thanks{$^{1}$Zach Dwiel is with Intel AI Labs, Bloomington, IN, U.S.A.
        {\tt\small zach.dwiel@intel.com}}%
\thanks{$^{2}$Madhavun Candadai is with the Program in Cognitive Science, and School of Informatics, Computing and Engineering, at Indiana University, Bloomington, IN, U.S.A.
        {\tt\small madcanda@indiana.edu}}%
\thanks{$^{3}$Mariano Phielipp is with Intel AI Labs, Phoenix, AZ, U.S.A.
        {\tt\small mariano.j.phielipp@intel.com}}%
}

\begin{document}

\maketitle
\thispagestyle{empty}
\pagestyle{empty}

%%%%%%%%%%%%%%%%%%%%%%%%%%%%%%%%%%%%%%%%%%%%%%%%%%%%%%%%%%%%%%%%%%%%%%%%%%%%%%%%
\begin{abstract}
The use of robotics in controlled environments has flourished over the last several decades and training robots to perform tasks using control strategies developed from dynamical models of their hardware have proven very effective. However, in many real-world settings, the uncertainties of the environment, the safety requirements and generalized capabilities that are expected of robots make rigid industrial robots unsuitable. This created great research interest into developing control strategies for flexible robot hardware for which building dynamical models are challenging. In this paper, inspired by the success of deep reinforcement learning (DRL) in other areas, we systematically study the efficacy of policy search methods using DRL in training flexible robots. Our results indicate that DRL is successfully able to learn efficient and robust policies for complex tasks at various degrees of flexibility. We also note that DRL using Deep Deterministic Policy Gradients can be sensitive to the choice of sensors and adding more informative sensors does not necessarily make the task easier to learn.
\end{abstract}

%%%%%%%%%%%%%%%%%%%%%%%%%%%%%%%%%%%%%%%%%%%%%%%%%%%%%%%%%%%%%%%%%%%%%%%%%%%%%%%%
\section{INTRODUCTION}
Inspired by the robustness and adaptability of natural systems, roboticists are becoming increasingly aware of the benefits of building soft and flexible robots~\cite{kim2013soft, rus2015design}.
While research in soft robots involving stretchable electronics is on one end of the spectrum, the other, perhaps more populated end, is dominated by research involving strict rigidity constraints on robot design.
%Both these extremes have their own advantages and limitations.
Fully soft robots could potentially mimic living systems and hence significantly outperform rigid robots, but pose many challenges in their fabrication and control~\cite{rus2015design}.
Rigid robots, on the other hand, are easy to control because simple models of their dynamics can be built, but are heavy and expensive due their strict compliance requirements and are constrained in the robot's adaptability~\cite{dwivedy2006dynamic}.
% fully-actuated rigid vs under-actuated rigid
Somewhere along this spectrum lies a class of robots that could be built from rigid hardware but can nevertheless allow for flexibility in their joints and material.
Research in this area involving the use of flexible actuators, has been of interest for several decades now and has shown great promise~\cite{benosman2004control, tokhi2008flexible}.
Reliable training methods for this class of robots would enable higher payload-to-weight ratios, higher speeds, lower cost, and safer operation due to less inertia~\cite{kiang2015review}.

With the aim of furthering this line of research, the primary questions that this paper addresses are the following:
\begin{enumerate}
    \item Can policy search serve as a reliable methodology for end-to-end training of flexible robots?
    \item Does learning with policy search enable generalization to levels of flexibility not seen during training?
    \item How sensitive are policy search methods to choice of sensors?
\end{enumerate}

% 1. Can policy search serve as a methodology for end-to-end training of flexible robots?\\
Deep artificial neural networks, in conjunction with reinforcement learning, namely deep reinforcement learning (DRL) has shown to be successful in a wide variety of problems~\cite{henderson2018deep}. As a learning methodology, a major advantage of the "deep" feature in DRL is that the task relevant features are also learnt and do not have to be hand-designed. Furthermore, DRL can learn complex tasks merely with reward signals as opposed to supervised learning where true labels are required. The presence of a reward function enables generation of easy training feedback signals removing the need for huge training datasets. In this paper we demonstrate that policy search using deep artificial neural networks on flexible hardware results in performance that is at least as good as (and sometimes better than) performance on rigid hardware.

% 2. Does learning with flexibility enable generalization to other levels of flexibility?\\
While learning and performance under flexibility is a desirable property in training algorithms, another important property is its ability to be robust to changes in flexibility. This is particularly the case when policies learned under specific levels of flexibility are transferred over to other levels of flexibility. An algorithm that is robust to different levels of flexibility places lower demands on the simulator to match the physical robot precisely or if training on real robots, lower demands are placed on the precision and consistency between robots. It also allows for sim2real transfer from one simulation to several robots that could vary in their flexibility, thus reducing constraints on robot hardware design to meet strict rigidity requirements. Furthermore, being robust to different levels of flexibility allows a policy to adapt to changes in robot dynamics due to age and environmental factors. In this paper we study the robustness of policies learned using policy search and demonstrate that they are in fact robust across a wide-range of flexibility levels.

% 3. How sensitive are policy search methods to choice of sensors?\\
Another aspect of designing flexible robots we study in this paper is the choice of sensors. Building models of robot dynamics often involve the assumption of a particular set of sensors, which then become part of the system of equations that define the model. With end-to-end training and policy search, there is no constraint on what sensors can be included and it adds minimal overhead to the robot design to add or remove sensors, in comparison to model based approaches. This can be an advantage because it is easy to try different sensor combinations, but it can also be a disadvantage because now sensor-choice is a hyper-parameter and the sensitivity of training to sensor-choice needs to be understood. In this paper, we demonstrate that policy search is sensitive to sensor choice and that adding sensors to provide more information to the learning algorithm is not necessarily helpful.

% The implications of these results are:\\
Being able to train flexible hardware using policy search provides several advantages to robot designers - it lowers the cost of robot hardware by dropping the rigidity constraint; complex hand-designed models of robot dynamics no longer need to be built, as policy search with DRL is a general purpose learning approach that enables end-to-end training; and it allows driving the same hardware at higher speeds and higher payloads.
Our demonstration of robustness of learned policy to flexibility implies resiliency across robots of different levels of flexibility as well as resiliency across physical changes over the course of the robot's lifetime.
This removes the need for periodically re-calibrating the model dynamics, with the model potentially getting increasingly complex with wear and tear of parts.
Ultimately, this approach takes us one step closer towards robots that are adaptive and robust like natural systems.

The rest of the paper is organized as follows: the next section talks about related work in this area, which is following by a section outlining the tasks and the specific policy search method used, which is then followed by sections detailing the experiments that were run, the results of those experiments, a discussion of the results and finally a conclusion.

\section{RELATED WORK}
Robot controller design is dominated by building precise mathematical models of its dynamics.
It is not always practical to build a general model of a robot's dynamics that is invariant to the various real-world factors ranging from noise to changes in the environment, motor backlash, motor torque output, or the focus of this paper, link flexibility.
In such cases, reinforcement learning and policy search algorithms that can learn from a robot's experience have been shown to be successful~\cite{deisenroth2013survey, kober2013reinforcement} for tasks such as object manipulation~\cite{gullapalli1995skillful, deisenroth2011learning, kalakrishnan2011learning}, locomotion~\cite{benbrahim1997biped, geng2006fast, endo2008learning, ha2018automated} and flight~\cite{ng2006autonomous}.
However, most of this work involves using a model-free component to approximate features of the robot or the world that cannot be modeled while still using model-based controllers for other parts of the system~\cite{kalakrishnan2011learning, liu2014adaptive}

%Mention reinforcement learning for policy search
In work where flexibility is taken into consideration, learning is still based either on building a more complex model~\cite{kiang2015review, chen2017hybrid, chen2018robust}, an approximate model~\cite{kumar2011reduced} or plugging in a learned-model component into a model-based controller.
Recently, work involving end-to-end model-free methods using deep reinforcement learning have been demonstrated successfully in rigid real robots~\cite{levine2016end, Gu2018Deep, ha2018automated}.
\cite{ha2018automated} have shown that learning directly in hardware is possible with policy search, they rightly point out that even in simple tasks, factors such as joint slackness etc. make it very difficult to train. While they and \cite{lillicrap2015continuous} have shown that this is possible using policy search, this paper systematically studies how policy search methods perform with flexible hardware.

%1 paragraph - sensor selection\\

%1 paragraph - information theory for robot design?\\

\section{METHODS}
\subsection{Robot design}
In order to systematically study the effect of flexibility on learning, we set up a single link robot arm on MuJoCo~\cite{todorov2012mujoco} in OpenAI gym like environments, with a ``pseudo-joint'' in the middle of the arm that the learner could not sense directly. The pseudo-joint was modeled with a spring joint so as to mimic a material bending from stress less than its proportionality limit, which is the stress limit under which the material stress strain curve follows Hooke's Law. We were able to scan through different levels of flexibility by adjusting the time-constant on the middle joint. The learning algorithm could only directly control the base joint and not the spring joint. The choice of sensors plays a crucial role in the amount of information that is available to the learner about the hardware flexibility. Intuitively, first order measures such as angle sensors would provide less information when compared to second or third order sensors such as accelerometers. Also, sensors placed on the tip of the arm and feet of the robots will be more sensitive to the flexing of the link than sensors on the joints. In order to study the ability of the policy search method to learn and potentially take advantage of flexibility we tried three different sensor configurations - observations that included inertial measurement units (IMU) along with angle sensors, observations that only had angle sensors and observations that only had the IMUs. Besides these sensors, task-specific inputs were provided depending on the nature of the task.

\subsection{Task design}
This section outlines the different tasks that the flexible robot arm was optimized to perform. We designed four tasks with increasing levels of difficulty based on the three canonical abilities that are typically expected of robot arms - navigating to a specific location, acting on objects in the environment and manipulating objects in the environment based on a goal. The ability of the policy search method to learn these canonical tasks under different levels of flexibility (including no flexibility) was tested for each task with different sensor configurations.

\subsubsection{Reacher}
The first task we tested on involved moving the tip of a flexible arm to a specific goal position by controlling the base joint torque (Fig.~\ref{fig:exp1}A). Besides the sensors specified above, the vector between the end of the robot arm and the goal position was added to the observations as in the Gym environment this task was derived from. Note that, due to its flexibility, the robot arm could be in many different configurations while observing the same goal vector. The robot arm was started at random positions within a range of $+/- 0.2$ radians from a reference position. The task was setup to incentivize reaching the goal as soon as possible, by providing a reward of -1 for every time step the arm was not at the goal and a reward of 0 when the it was within a euclidean threshold distance of $0.05$ of the goal. For reference, the length of the arm is $0.2$. Besides the distance based reward, there was an energy penalty in each time step that was estimated based on the motor action as $0.1 * action^2$. A training episode lasted until the goal was reached or for 200 time steps, whichever was earlier.

\subsubsection{Reacher Stay}
This task was identical to `Reacher` except that in this case, the robot arm was expected to stay at the goal position upon reaching it (Fig.~\ref{fig:exp1}B). This was achieved by not finishing a training episode upon reaching the goal and instead providing a reward of 0 for staying at the goal position, and again a reward of -1 for each step it is not at the goal position. An episode in this case lasted for a fixed duration of 200 time steps during which reward can be maximized by reaching the goal position as quickly as possible and staying there. Note that in flexible robots, efficiently performing this task requires that vibrations are damped so as to arrive at a stop in the goal position and stay there.

\subsubsection{Thrower}
Flexibility in a robot arm could either deter or aid its ability to interact with an object in the environment. In this task, the goal for the robot arm was to ``throw'' a puck as far as possible (Fig.~\ref{fig:exp1}C). The puck is placed at a fixed position in the robot's environment and the arm starts at random positions similar to the previous tasks. Also similar to previous tasks, the observation includes the vector between the fingertip of the robot and the object position. Based on its flexibility, the robot should learn to push the object appropriately so as to generate maximum acceleration of the puck. In this task, the only reward the robot receives during an episode is the energy penalty, but at the end of 200 time steps, a reward proportional to the distance between the initial and final position of the puck is provided.

\subsubsection{Ant}
While the tasks discussed so far involved a single link robot, with a single flexible component, in order to test more complex tasks we modified the Ant environment in Open AI gym. This task involved training a four-legged ant to walk with rewards being provided at each time step proportional to the distance walked, the energy spent, contact with the floor and for maintaining stable dynamics (Fig.~\ref{fig:exp1}D). The Ant was made flexible by the addition of two flexible ball joints to each leg - one between the hip and the knee, and other between the knee and the ankle. Also unlike previous tasks, these pseudo-joints, being ball joints, have higher degree of freedom to be flexible, and since all four legs have these joints, learning to walk involves the coordination of multiple flexible parts, thus making this task significantly more difficult to learn.

\subsection{Policy Search Method: DDPG}
Deep Deterministic Policy Gradients (DDPG) is a stochastic reward based policy search method that has been shown to stably learn policies for continuous observation and action spaces~\cite{lillicrap2015continuous}. The architecture involves an actor-critic neural network pair, where the actor takes in observations to provide actions that can be performed, and the critic provides an estimate of the expected discounted long-term reward given the policy. The critic is trained to better estimate the expected long-term reward based on the actual rewards received from the environment. The actor is trained using the training signal provided by the critic to adjust its parameters in the direction that maximizes expected long-term reward. The training dynamics between these two networks is stabilized by having copies of these networks that are updated at a slower time-scale, and whose targets essentially convert the training of the main actor and critic to that of a supervised learning problem. In~\cite{lillicrap2015continuous} it has been demonstrated that DDPG can efficiently learn competitive policies across a wide range of tasks. Hyper-parameters such as network architecture (2 hidden layers with 400 and 300 neurons respectively, identical actor and critic networks), with minibatch sizes of 64, replay buffer size of $10^6$, learning rate ($10^{-4}$ and $10^{-3}$ for actor and critic respectively), were same as the original DDPG paper and did not require any fine tuning to for each task. Intel's {\em RL-Coach}~\cite{caspi_itai_2017_1134899} implementation of DDPG was employed in conjunction with customized MuJoCo environments to perform the experiments outlined in the next section.

\subsection{Information theoretic analysis}
In the analysis of the impact of different sensor choices on learning performance, we measure the mutual information between the observations and the end-effector position while running a random policy and similarly between the observations and rewards while running a trained policy. This was carried out using the python {\em infotheory} package~\cite{candadai2019infotheory}, where data distributions were estimated using equal interval binning followed by average shifted-histograms~\cite{scott1985averaged} in order to smoothen the boundary effects of arbitrary binning. Two different binning resolutions, 25 and 50, were tried giving similar results. In both cases, shifted histograms with 3 shifts was used to estimate the final data distribution. Results reported in the paper are from binning with 50 bins along each dimension.

\begin{figure*}[t]
    \centering
    \includegraphics[width=\textwidth]{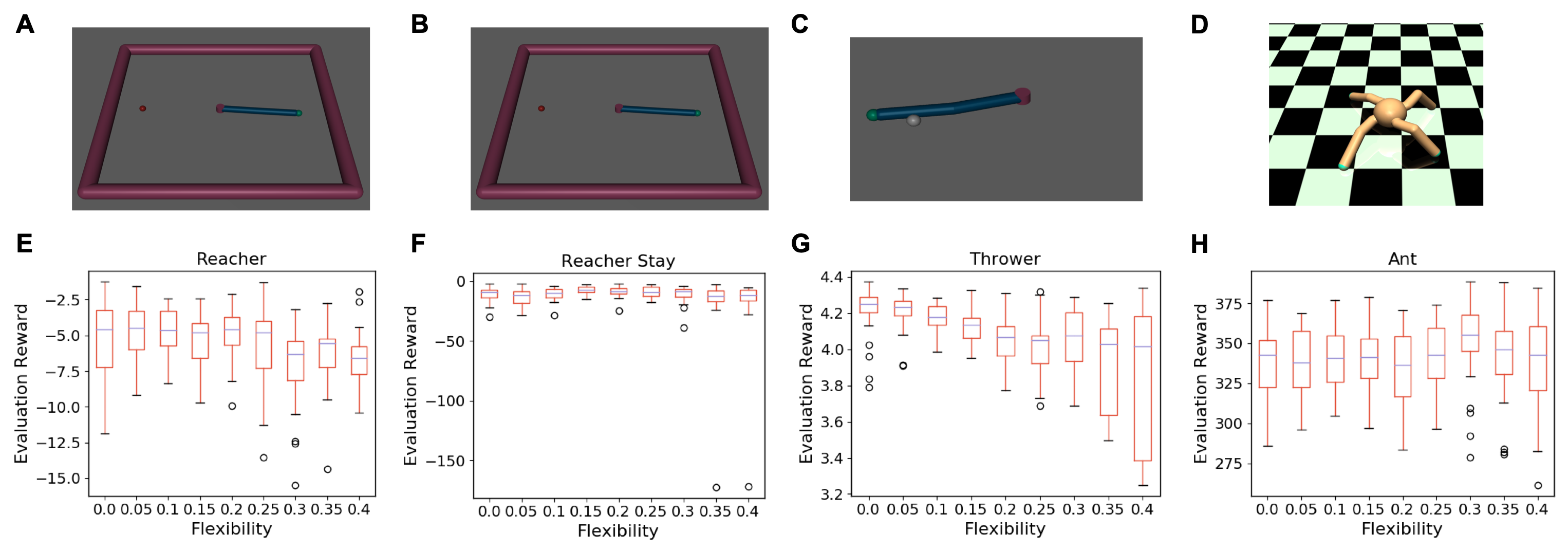}
    \caption{Flexible robots, tasks and performance. [A-D] The four tasks outlined in the methods section - Reacher, Reacher Stay, Thrower and Ant. The Reacher and Reacher stay tasks have a similar setup. [E-H] Performance of DDPG on the four tasks as a function of robot flexibility from 20 runs for each level of flexibility. These plots show that with joint angle sensors alone, DDPG is able to perform consistently well across all levels of flexibility.}
    \label{fig:exp1}
\end{figure*}

\begin{figure*}
    \centering
    \includegraphics[width=\textwidth]{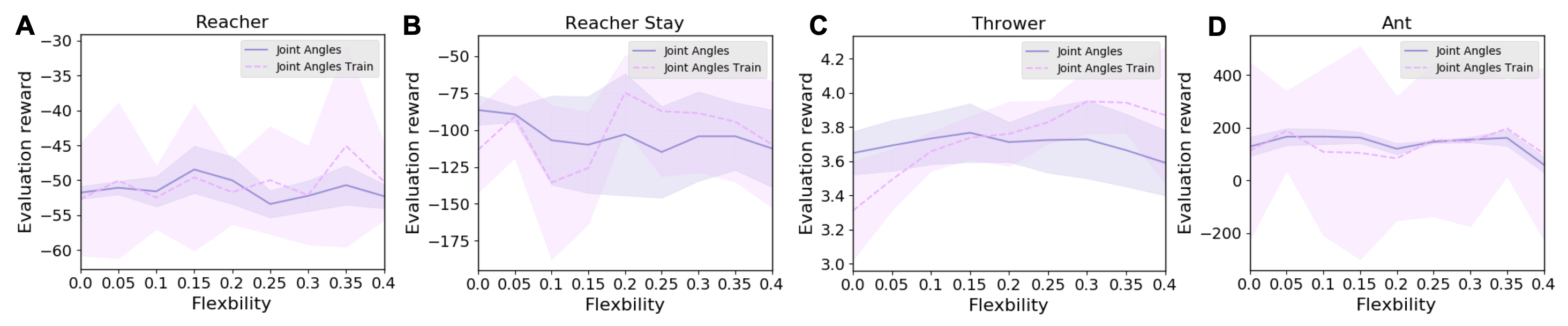}
    \caption{Performance of policy search on levels of flexibility it was not trained on. Across the four tasks, solid lines represent mean performance (across 20 runs) in the task the policy was trained in with one standard deviation shaded on either side, and the dashed lines represent mean performance of the same policy in all other flexibility levels (mean of mean total reward across 100 episodes for each level of flexibility) that it was not trained on. DDPG performs reliably across tasks and is robust in its performance across different levels of flexibility.}
    \label{fig:robustness}
\end{figure*}

\section{EXPERIMENTS}
The methods section outlined the robot and task design, the policy search method and the analysis techniques. This section outlines the specific experiments that were conducted to systematically study how flexibility affects learning performance of policy search.

\subsection{Can policy search serve as a methodology for end-to-end training of flexible robots?}
The first experiment involved studying if policy search methods, as an end-to-end learning methodology, can learn to perform tasks when hardware components were flexible. In order to study this, we optimized deep neural network policies using DPPG for the tasks mentioned in the previous section while systematically increasing the level of flexibility. For each task, for each level of flexibility, 20 independent runs with different random seeds were executed to study the variance in performance.

\subsection{Are policy search methods robust to other levels of flexibility?}
The second experiment involved studying policies trained on one level of flexibility, and tested in all other levels of flexibility. The focus of this experiment is to study the relative change in performance when tested on levels flexibility other than level it was trained on. An algorithm can be considered robust if it performs comparably well with perturbations during testing in relation to its performance during training. To this end, of the 9 different levels of flexibility that was available, we evaluated policies optimized for one level of flexibility to its performance on the remaining 8 levels. This was repeated for each of the 20 runs and for each level of flexibility and task.

\subsection{Are policy search methods sensitive to choice of sensors?}
Next, we studied the effect choice of sensors had in training flexible robots using policy search by including IMU sensors in the robot in addition to joint angle sensors. The intuition being, in flexible robots, the higher order IMU sensor provides more information about the flexible dynamics than first-order joint angle sensors. Training was carried out on the same tasks with IMUs added at each end-effector, in addition to the joint angle sensors. Similar to experiment 1, the maximum mean evaluation reward from 100 evaluation trials across training time was considered the representative solution, and this was repeated for 20 runs for each sensor choice and for each level of flexibility. Performance-flexibility curves were compared across the different sensor choices.

\begin{figure*}
    \centering
    \includegraphics[width=\textwidth]{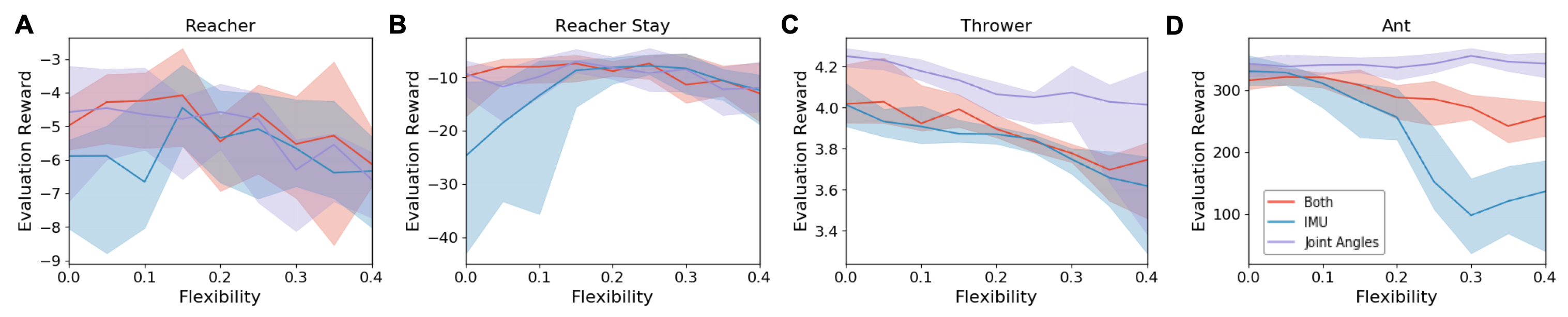}
    \caption{Performance of policy search with different sensor choices. Across the four tasks, DDPG performs well in across all levels of flexibility only when joint angle sensors alone are used (purple line in all panels). With IMUs only or with both sensors, it is at best the same (Reacher and Reacher Stay) or worse (Thrower and Ant)}
    \label{fig:sensor_choices}
\end{figure*}

\begin{figure*}
    \centering
    \includegraphics[width=0.8\textwidth]{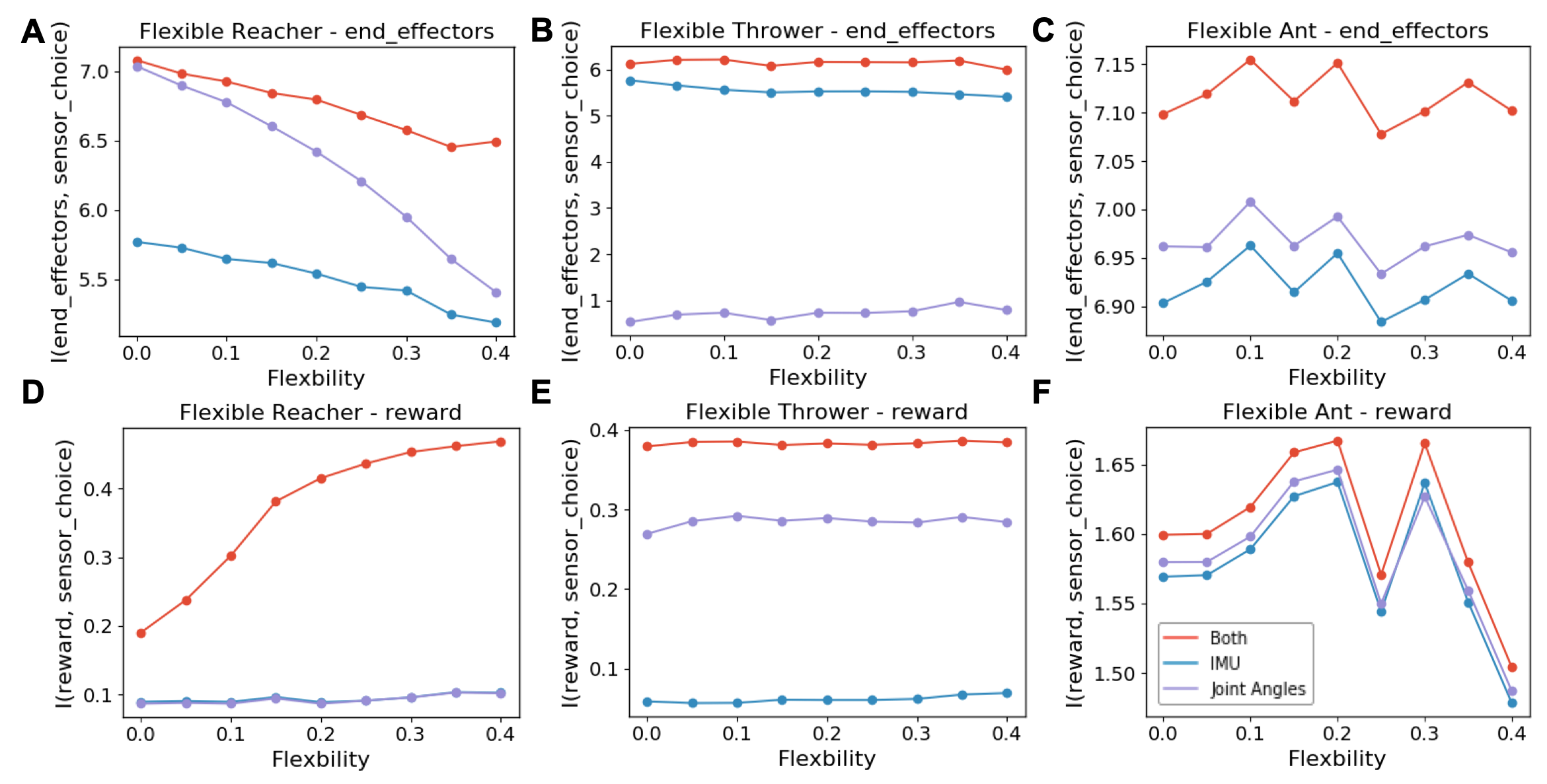}
    \caption{Informational analysis of sensor choice from a random policy. Top row: Information about position of the end-effector in each sensor choice for the three robot configurations (Reacher and Reacher Stay are the same) Bottom row: Information about reward in each sensor choice for the corresponding robot configurations as top row.}
    \label{fig:info}
\end{figure*}

\section{RESULTS}
\subsection{Policy search is resilient to flexibility of robot hardware}
As is common with DDPG, the policy that performed the best during the 100 evaluation episodes was selected as the representative solution for a particular run. Comparing the mean and standard deviations of the best policies over 20 independent runs, showed that DDPG was able to reliably learn efficient policies across all levels of flexibility tested using joint angle sensors alone as observations~\ref{fig:exp1}.
While Reacher and Thrower showed a gradual, albeit small, decline in performance with increasing levels of flexibility, the Reacher Stay and Ant tasks performed consistently well across all levels of flexibility.
This demonstrates that while end-to-end learning in flexible robots with DDPG can learn perform efficiently with flexible hardware in all tasks, it can also learn to take advantage of flexibility in robot hardware in order to perform even better.

\subsection{Policies learnt were robust in adapting to different levels of flexibility}
The second experiment that was performed involved studying the robustness of policies learned using DDPG across different levels of flexibility. The average evaluation reward over 100 episodes was estimated for each level of flexibility that the robot was not trained on, and compared to the performance on the level of flexibility it was trained on.
Policies from the end of training in all 4 tasks show that policies learned using DDPG were able to adapt robustly and earn a total reward similar to the reward received on the task they were trained on~\ref{fig:robustness}.
In other words, the performance is invariant to the flexibility of the robot thereby making DDPG a robust approach to training flexible robots.

\subsection{Policy search was sensitive to sensor choice.}
Policy search performs at least as well if not better with joint angles, when compared to performance with the addition of IMU sensors. Contrary to intuition, adding the additional sensor did not result in better performance. In the Reacher and Reacher Stay tasks, addition of IMU results in performance that was just as good, but in the Thrower and Ant tasks, it resulted in worse performance.
DDPG performs with no difference between sensor choices in the Reacher task~(\ref{fig:sensor_choices}A), suffers some drop in performance of Reacher Stay with IMUs only~(\ref{fig:sensor_choices}B), drops in performance with IMUs only and both sensors in Thrower~(\ref{fig:sensor_choices}C), and finally drop significantly with flexibility for Ant with IMUs only and also with both sensors~(\ref{fig:sensor_choices}D). Joint angles alone is the only sensor choice for which DDPG performs consistently well across all levels of flexibility.
These results demonstrate that policy search is sensitive to the choice of sensors. This is explored further next.

\subsection{More information with more sensors is not indicative of better performance.}
While the first two results provide general support for the use of policy search in training flexible robots, the third result above, which shows DDPG to be quite susceptible to sensor choice is counter-intuitive. In order to better understand where intuitions about training with more sensors may be wrong, we analyzed if the addition of IMU sensors did in fact provide more information about the task by measuring, in a random policy, the amount of mutual information between (a) sensor observations and position of the end-effector and (b) sensor observations and reward. Note that, when the robot arm is flexible, joint angle sensors at the base would not accurately determine the position of the end-effector.
Performing these analyses on the three robot settings revealed that adding IMU sensors does increases information about end-effector position thereby providing the agent with more information about the flexible dynamics of the robot in its observation~(Top row in Fig.~\ref{fig:info}).
Furthermore, the additional information was shown to be not just about robot dynamics but also in relevance to the task, because information about reward in the observations also consistently went up in all tasks with the addition of IMU sensors~(bottom row in Fig.~\ref{fig:info}).

Interestingly, in Reacher, angle sensors provided more information than IMU about position of end-effector~(Fig.~\ref{fig:info}A), but together they provided significantly more information about the reward that either one of them alone~(Fig.~\ref{fig:info}D) which suggests the combination should perform really well in extreme levels of flexibility.
With Thrower, most likely due to the greater impact of flexibility when the arm comes into contact with an object in the environment, IMU gives significantly more information about object position~(Fig.~\ref{fig:info}B). While striking the object, the base angle of the arm could be in a variety of positions that would have never occurred in the Reacher tasks. This results in the angle sensor having very low information about the position of the end-effector. However, joint angle sensors show more information about the reward than the IMUs(Fig.~\ref{fig:info}E). This suggests that although addition of a sensor provides more information about the dynamics of the arm, that information is not quite relevant to the task, although together it does contribute to the reward to some degree.
In Ant, addition of IMUs show that it did not add much to the magnitude of the information about the end-effector or reward~(Fig.~\ref{fig:info}C and F), while the IMU itself provided a lot of information which suggests that the two sensors provide redundant information in different formats that the network might have to learn to parse.

Intuitive understanding about the role of additional sensors is composed of two parts - the first being, addition of more sensors should provide more information to learn from, and the second being, having access to more information should make learning easier.
DDPG's performance suggests that, in line with the first part of our intuition, addition of sensors does increase the amount of task-relevant information to the learner, but it is the second part that does not appear to hold. More information does not necessarily make it easier for policy learning. This could be further explored in future studies involving controlled experiments of systematic manipulation of information in different tasks to better understand the relationship between information and learning in deep networks using reinforcement learning.

\section{DISCUSSION}
In summary, through the experiments with Deep Deterministic Policy Gradients as a policy search method for training flexible robots, we have demonstrated the following (i) policy search methods are capable of end-to-end learning to perform efficiently across a wide range of flexibility in hardware (ii) policies learned using DDPG are robust in their ability to adapt to different levels of flexibility than the one they were trained on and (iii) while policy search methods learn and perform well with the correct choice of sensors, they can be susceptible to the choice of sensor. Thus, in the training of flexible robots using policy search, appropriate sensor choice is the more crucial parameter than the flexibility itself.

The ability of learned policies to be robust across different levels of hardware flexibility offers several advantages as noted previously. Besides simply dropping the constraint on rigidity requirements of the hardware, this property of learned policies also allows transfer from rigid simulators to flexible robots or from flexible robots to other differently flexible robots. Furthermore, they also allow transfer of policies between robots that were manufactured with variance in properties from one robot to another. Thus, taking this approach decreases constraints and increases generalizability in all aspects of robot design.

The work in this paper has taken a systematic approach to studying the properties of policy search methods for training flexible robots. While we have tested on a relatively new algorithm, DDPG, there are certainly other even newer algorithms such as TRPO, PPO and SAC that could be tested. Perhaps they are less susceptible to sensor choices. Future work in our group involves studying more algorithms in a greater number of tasks, learning directly in real flexible robots as opposed to simulation, and further understanding the relationship between input information to a learner and its ability to learn.

This paper demonstrates, as a proof-of-concept, the efficacy and sensitivities of using policy search for end-to-end training of flexible robots. Artificial neural networks, being universal function approximators and reinforcement learning being a very practical training approach for robots form a potent combination that simplify robot design by relaxing one of the most widely imparted constraint on robots - rigidity. This would make robots cheaper, easier to design and maintain, and more robust in the face of changes to robot dynamics that would otherwise require a complete rebuilding of its model from scratch. While there are still several aspects of this approach that are yet to be explored, such as the relationship between information content in the observations and the ability to learn, this paper demonstrates that policy search holds a lot of promise for robot design that could get closer to that of natural systems.

%\addtolength{\textheight}{-12cm}   % This command serves to balance the column lengths
                                  % on the last page of the document manually. It shortens
                                  % the textheight of the last page by a suitable amount.
                                  % This command does not take effect until the next page
                                  % so it should come on the page before the last. Make
                                  % sure that you do not shorten the textheight too much.

%%%%%%%%%%%%%%%%%%%%%%%%%%%%%%%%%%%%%%%%%%%%%%%%%%%%%%%%%%%%%%%%%%%%%%%%%%%%%%%%

%%%%%%%%%%%%%%%%%%%%%%%%%%%%%%%%%%%%%%%%%%%%%%%%%%%%%%%%%%%%%%%%%%%%%%%%%%%%%%%%

%%%%%%%%%%%%%%%%%%%%%%%%%%%%%%%%%%%%%%%%%%%%%%%%%%%%%%%%%%%%%%%%%%%%%%%%%%%%%%%%
%\section*{APPENDIX}

%Appendixes should appear before the acknowledgment.

%\section*{ACKNOWLEDGMENT}

%The preferred spelling of the word �acknowledgment� in America is without an �e� after the �g�. Avoid the stilted expression, �One of us (R. B. G.) thanks . . .�  Instead, try �R. B. G. thanks�. Put sponsor acknowledgments in the unnumbered footnote on the first page.

%%%%%%%%%%%%%%%%%%%%%%%%%%%%%%%%%%%%%%%%%%%%%%%%%%%%%%%%%%%%%%%%%%%%%%%%%%%%%%%%

\bibliographystyle{IEEEtran}
\bibliography{IEEEexample}

\begin{thebibliography}{10}
\providecommand{\url}[1]{#1}
\csname url@rmstyle\endcsname
\providecommand{\newblock}{\relax}
\providecommand{\bibinfo}[2]{#2}
\providecommand\BIBentrySTDinterwordspacing{\spaceskip=0pt\relax}
\providecommand\BIBentryALTinterwordstretchfactor{4}
\providecommand\BIBentryALTinterwordspacing{\spaceskip=\fontdimen2\font plus
\BIBentryALTinterwordstretchfactor\fontdimen3\font minus
  \fontdimen4\font\relax}
\providecommand\BIBforeignlanguage[2]{{%
\expandafter\ifx\csname l@#1\endcsname\relax
\typeout{** WARNING: IEEEtran.bst: No hyphenation pattern has been}%
\typeout{** loaded for the language `#1'. Using the pattern for}%
\typeout{** the default language instead.}%
\else
\language=\csname l@#1\endcsname
\fi
#2}}

\bibitem{kim2013soft}
S.~Kim, C.~Laschi, and B.~Trimmer, ``Soft robotics: a bioinspired evolution in
  robotics,'' \emph{Trends in biotechnology}, vol.~31, no.~5, pp. 287--294,
  2013.

\bibitem{rus2015design}
D.~Rus and M.~T. Tolley, ``Design, fabrication and control of soft robots,''
  \emph{Nature}, vol. 521, no. 7553, p. 467, 2015.

\bibitem{dwivedy2006dynamic}
S.~K. Dwivedy and P.~Eberhard, ``Dynamic analysis of flexible manipulators, a
  literature review,'' \emph{Mechanism and machine theory}, vol.~41, no.~7, pp.
  749--777, 2006.

\bibitem{benosman2004control}
M.~Benosman and G.~Le~Vey, ``Control of flexible manipulators: A survey,''
  \emph{Robotica}, vol.~22, no.~5, pp. 533--545, 2004.

\bibitem{tokhi2008flexible}
M.~O. Tokhi and A.~K. Azad, \emph{Flexible robot manipulators: modelling,
  simulation and control}.\hskip 1em plus 0.5em minus 0.4em\relax Iet, 2008,
  vol.~68.

\bibitem{kiang2015review}
C.~T. Kiang, A.~Spowage, and C.~K. Yoong, ``Review of control and sensor system
  of flexible manipulator,'' \emph{Journal of Intelligent \& Robotic Systems},
  vol.~77, no.~1, pp. 187--213, 2015.

\bibitem{henderson2018deep}
P.~Henderson, R.~Islam, P.~Bachman, J.~Pineau, D.~Precup, and D.~Meger, ``Deep
  reinforcement learning that matters,'' in \emph{Thirty-Second AAAI Conference
  on Artificial Intelligence}, 2018.

\bibitem{deisenroth2013survey}
M.~P. Deisenroth, G.~Neumann, J.~Peters, \emph{et~al.}, ``A survey on policy
  search for robotics,'' \emph{Foundations and Trends{\textregistered} in
  Robotics}, vol.~2, no. 1--2, pp. 1--142, 2013.

\bibitem{kober2013reinforcement}
J.~Kober, J.~A. Bagnell, and J.~Peters, ``Reinforcement learning in robotics: A
  survey,'' \emph{The International Journal of Robotics Research}, vol.~32,
  no.~11, pp. 1238--1274, 2013.

\bibitem{gullapalli1995skillful}
V.~Gullapalli, ``Skillful control under uncertainty via direct reinforcement
  learning,'' \emph{Robotics and autonomous systems}, vol.~15, no.~4, pp.
  237--246, 1995.

\bibitem{deisenroth2011learning}
M.~P. Deisenroth, C.~E. Rasmussen, and D.~Fox, ``Learning to control a low-cost
  manipulator using data-efficient reinforcement learning,'' 2011.

\bibitem{kalakrishnan2011learning}
M.~Kalakrishnan, L.~Righetti, P.~Pastor, and S.~Schaal, ``Learning force
  control policies for compliant manipulation,'' in \emph{Intelligent Robots
  and Systems (IROS), 2011 IEEE/RSJ International Conference on}.\hskip 1em
  plus 0.5em minus 0.4em\relax IEEE, 2011, pp. 4639--4644.

\bibitem{benbrahim1997biped}
H.~Benbrahim and J.~A. Franklin, ``Biped dynamic walking using reinforcement
  learning,'' \emph{Robotics and Autonomous Systems}, vol.~22, no. 3-4, pp.
  283--302, 1997.

\bibitem{geng2006fast}
T.~Geng, B.~Porr, and F.~W{\"o}rg{\"o}tter, ``Fast biped walking with a
  reflexive controller and real-time policy searching,'' in \emph{Advances in
  Neural Information Processing Systems}, 2006, pp. 427--434.

\bibitem{endo2008learning}
G.~Endo, J.~Morimoto, T.~Matsubara, J.~Nakanishi, and G.~Cheng, ``Learning
  cpg-based biped locomotion with a policy gradient method: Application to a
  humanoid robot,'' \emph{The International Journal of Robotics Research},
  vol.~27, no.~2, pp. 213--228, 2008.

\bibitem{ha2018automated}
S.~Ha, J.~Kim, and K.~Yamane, ``Automated deep reinforcement learning
  environment for hardware of a modular legged robot,'' in \emph{2018 15th
  International Conference on Ubiquitous Robots (UR)}.\hskip 1em plus 0.5em
  minus 0.4em\relax IEEE, 2018, pp. 348--354.

\bibitem{ng2006autonomous}
A.~Y. Ng, A.~Coates, M.~Diel, V.~Ganapathi, J.~Schulte, B.~Tse, E.~Berger, and
  E.~Liang, ``Autonomous inverted helicopter flight via reinforcement
  learning,'' in \emph{Experimental Robotics IX}.\hskip 1em plus 0.5em minus
  0.4em\relax Springer, 2006, pp. 363--372.

\bibitem{liu2014adaptive}
H.~Liu and T.~Zhang, ``Adaptive neural network finite-time control for
  uncertain robotic manipulators,'' \emph{Journal of Intelligent \& Robotic
  Systems}, vol.~75, no. 3-4, pp. 363--377, 2014.

\bibitem{chen2017hybrid}
Z.~Chen, T.~Zhang, and Z.~Li, ``Hybrid control scheme consisting of adaptive
  and optimal controllers for flexible-base flexible-joint space manipulator
  with uncertain parameters,'' in \emph{Intelligent Human-Machine Systems and
  Cybernetics (IHMSC), 2017 9th International Conference on}, vol.~1.\hskip 1em
  plus 0.5em minus 0.4em\relax IEEE, 2017, pp. 341--345.

\bibitem{chen2018robust}
T.~Zhang and Z.~Li, ``Robust-adaptive neural network finite-time control and
  vibration suppression for flexible-base flexible-joint space manipulator,''
  in \emph{2018 10th International Conference on Intelligent Human-Machine
  Systems and Cybernetics (IHMSC)}, vol.~1.\hskip 1em plus 0.5em minus
  0.4em\relax IEEE, 2018, pp. 224--229.

\bibitem{kumar2011reduced}
A.~Kumar, P.~M. Pathak, and N.~Sukavanam, ``Reduced model based control of two
  link flexible space robot,'' \emph{Intelligent control and automation},
  vol.~2, no.~02, p. 112, 2011.

\bibitem{levine2016end}
S.~Levine, C.~Finn, T.~Darrell, and P.~Abbeel, ``End-to-end training of deep
  visuomotor policies,'' \emph{The Journal of Machine Learning Research},
  vol.~17, no.~1, pp. 1334--1373, 2016.

\bibitem{Gu2018Deep}
\BIBentryALTinterwordspacing
S.~Gu, E.~Holly, T.~Lillicrap, and S.~Levine, ``Deep reinforcement learning for
  robotic manipulation with asynchronous off-policy updates,'' 2017. [Online].
  Available: \url{https://arxiv.org/abs/1610.00633}
\BIBentrySTDinterwordspacing

\bibitem{lillicrap2015continuous}
T.~P. Lillicrap, J.~J. Hunt, A.~Pritzel, N.~Heess, T.~Erez, Y.~Tassa,
  D.~Silver, and D.~Wierstra, ``Continuous control with deep reinforcement
  learning,'' \emph{arXiv preprint arXiv:1509.02971}, 2015.

\bibitem{todorov2012mujoco}
E.~Todorov, T.~Erez, and Y.~Tassa, ``Mujoco: A physics engine for model-based
  control,'' in \emph{2012 IEEE/RSJ International Conference on Intelligent
  Robots and Systems}.\hskip 1em plus 0.5em minus 0.4em\relax IEEE, 2012, pp.
  5026--5033.

\bibitem{caspi_itai_2017_1134899}
\BIBentryALTinterwordspacing
I.~Caspi, G.~Leibovich, G.~Novik, and S.~Endrawis, ``Reinforcement learning
  coach,'' Dec. 2017. [Online]. Available:
  \url{https://doi.org/10.5281/zenodo.1134899}
\BIBentrySTDinterwordspacing

\bibitem{candadai2019infotheory}
M.~Candadai and E.~J. Izquierdo, ``infotheory: A c++/python package for
  multivariate information theoretic analysis,'' \emph{arXiv preprint
  arXiv:1907.02339}, 2019.

\bibitem{scott1985averaged}
D.~W. Scott, ``Averaged shifted histograms: effective nonparametric density
  estimators in several dimensions,'' \emph{The Annals of Statistics}, pp.
  1024--1040, 1985.

\end{thebibliography}

\end{document}